\documentclass[11pt,a4paper]{article}
\usepackage{authblk}
\usepackage[hyperref]{acl2018}
\usepackage{times}
\usepackage{latexsym}
\usepackage{amsmath}
\usepackage{enumitem}
\usepackage{multirow}
\usepackage{multicol}
\usepackage{subcaption}
\usepackage{amssymb}
\usepackage[pdftex]{graphicx}

\allowdisplaybreaks
\DeclareMathOperator*{\softmax}{softmax}
\graphicspath{{./figures/}}
\usepackage{url}

\aclfinalcopy 
\def\aclpaperid{1570} 

\renewcommand{\footnotesize}{\fontsize{8pt}{10pt}\selectfont}

\title{A Walk-based Model on Entity Graphs for Relation Extraction}

\author[1,3]{Fenia Christopoulou}
\author[2,3]{Makoto Miwa}
\author[1,3]{Sophia Ananiadou}
\affil[1]{National Centre for Text Mining}
\affil[1]{School of Computer Science, The University of Manchester, United Kingdom}
\affil[2]{Toyota Technological Institute, Nagoya, 468-8511, Japan}
\affil[3]{Artificial Intelligence Research Center (AIRC), \authorcr
\textnormal{\normalsize National Institute of Advanced Industrial Science and Technology (AIST), Japan}}
\affil[ ]{\tt \{efstathia.christopoulou, sophia.ananiadou\}@manchester.ac.uk}
\affil[ ]{\tt makoto-miwa@toyota-ti.ac.jp}

\date{}

\begin{document}
	\maketitle
    
	\begin{abstract}
		We present a novel graph-based neural network model for relation extraction. 
        Our model treats multiple pairs in a sentence simultaneously and considers interactions among them. 
        All the entities in a sentence are placed as nodes in a fully-connected graph structure. The edges are represented with position-aware contexts around the entity pairs. 
		In order to consider different relation paths between two entities, we construct up to $l$-length walks between each pair.
		The resulting walks are merged and iteratively used to update the edge representations into longer walks representations. 
		We show that the model achieves performance comparable to the state-of-the-art systems on the ACE 2005 dataset without using any external tools\footnote{Source code available at \url{https://github.com/fenchri/walk-based-re}}. 
	\end{abstract}

	\section{Introduction}
    
\begin{figure}[t!]
		\centering
		\includegraphics[scale=0.53]{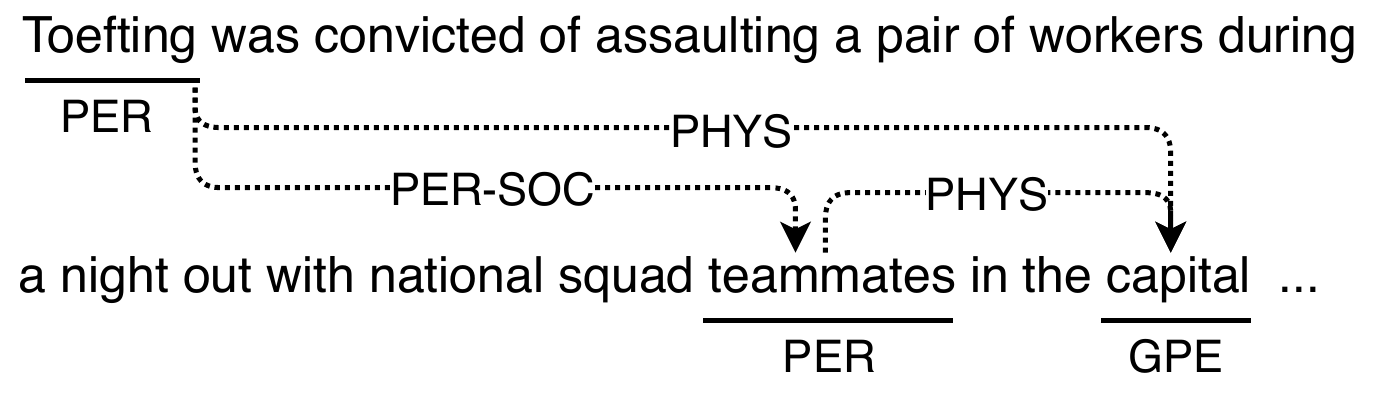}
		\caption{Relation examples from ACE (Automatic Content Extraction) 2005 dataset~\cite{doddington2004automatic}.}
		\label{fig:ex}
	\end{figure}

	Relation extraction (RE) is a task of identifying typed relations between known entity mentions in a sentence. 
    Most existing RE models treat each relation in a sentence individually~\cite{miwa2016end,nguyen2015perspective}.
    However, a sentence typically contains multiple relations between entity mentions.
    RE models need to consider these pairs simultaneously to model the dependencies among them.
    The relation between a pair of interest (namely ``target" pair) can be influenced by other pairs in the same sentence. 
	The example illustrated in Figure~\ref{fig:ex} explains this phenomenon. 
	The relation between the pair of interest \textit{Toefting} and \textit{capital}, can be extracted directly from the target entities or indirectly by incorporating information from other related pairs in the sentence. 
	The person entity (PER) \textit{Toefting} is directly related with \textit{teammates} through the preposition \textit{with}.
	Similarly, \textit{teammates} is directly related with the geopolitical entity (GPE) \textit{capital} through the preposition \textit{in}.
    \textit{Toefting} and \textit{capital} can be directly related through \textit{in} or indirectly related through \textit{teammates}.
	Substantially, the path from \textit{Toefting} to \textit{teammates} to \textit{capital} can additionally support the relation between \textit{Toefting} and \textit{capital}.
    
	Multiple relations in a sentence between entity mentions can be represented as a graph.
    Neural graph-based models have shown significant improvement in modelling graphs 
    over traditional feature-based approaches in several tasks. 
    They are most commonly applied on knowledge graphs (KG) for knowledge graph completion~\cite{jiang2017attentive} and the creation of knowledge graph embeddings~\cite{wang2017knowledge,shi2017proje}. 
    These models rely on paths between existing relations in order to infer new associations between entities in KGs. 
	However, for relation extraction from a sentence, related pairs are not predefined and consequently all entity pairs need to be considered to extract relations. 
    In addition, state-of-the-art RE models sometimes depend on external syntactic tools to build the shortest dependency path (SDP) between two entities in a sentence~\cite{xu2015neg,miwa2016end}. 
    This dependence on external tools leads to domain dependent models. 
    
	In this study, we propose a neural relation extraction model based on an entity graph, where entity mentions constitute the nodes and directed edges correspond to ordered pairs of entity mentions.
    The overview of the model is shown in Figure~\ref{fig:model}. 
    We initialize the representation of an edge (an ordered pair of entity mentions) from the representations of the entity mentions and their context.
    The context representation is achieved by employing an attention mechanism on context words.
    We then use an iterative process to aggregate up-to $l$-length walk representations between two entities into a single representation, which corresponds to the final representation of the edge.
    
    The contributions of our model can be summarized as follows:
	\begin{itemize}[nolistsep]
		\item We propose a graph walk based neural model that considers multiple entity pairs in relation extraction from a sentence.
		\item We propose an iterative algorithm to form a single representation for up-to $l$-length walks between the entities of a pair.
		\item We show that our model performs comparably to the state-of-the-art without the use of external syntactic tools. 
	\end{itemize}

	\section{Proposed Walk-based Model}
 
    The goal of the RE task is given a sentence, entity mentions and their semantic types, to extract and classify all related entity pairs (target pairs) in the sentence.
	The proposed model consists of five stacked layers: embedding layer, BLSTM Layer, edge representation layer, walk aggregation layer and finally a classification layer. 
	
\begin{figure}[t!]
		\includegraphics[width=\linewidth]{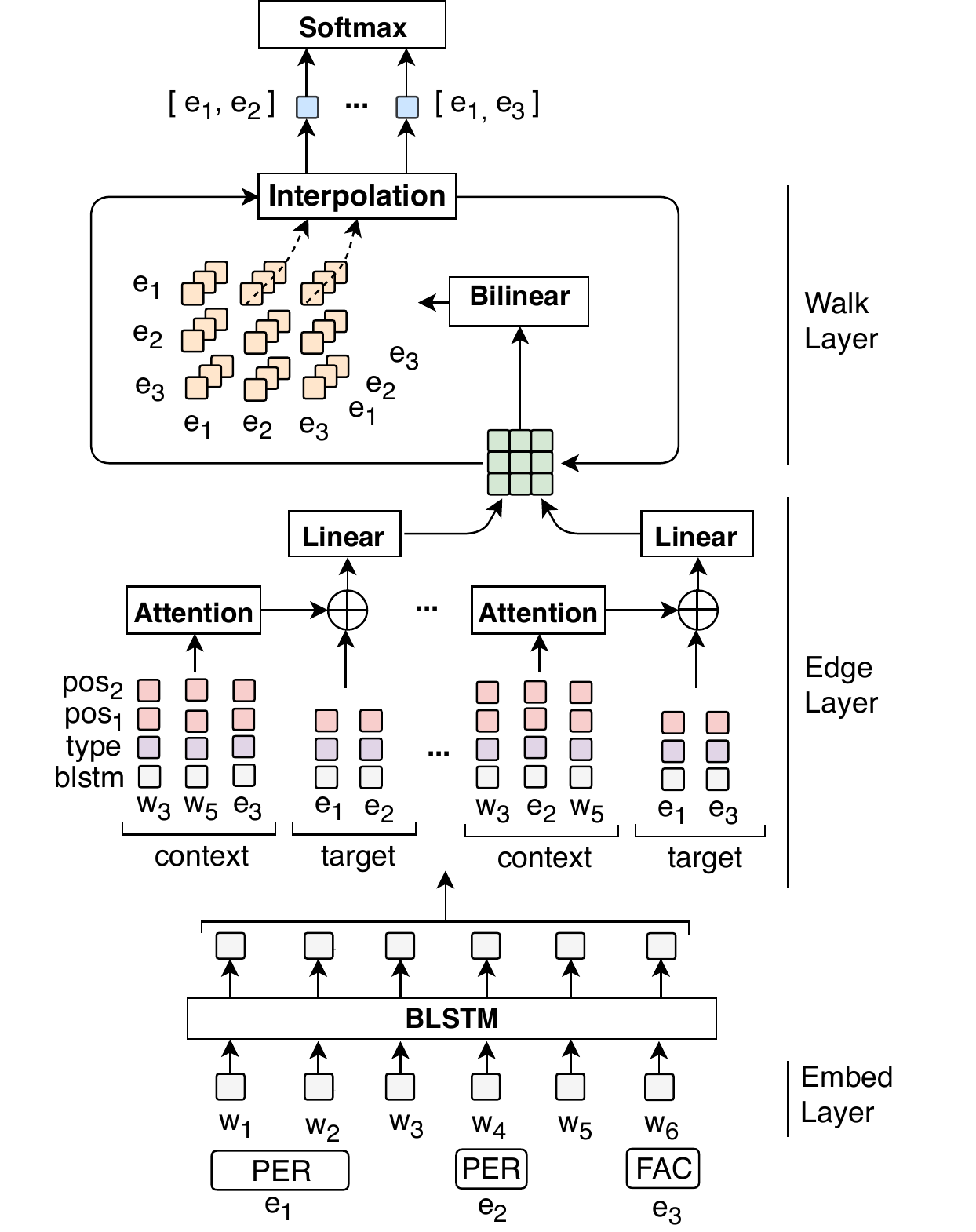}
		\caption{Overview of the walk-based model.}
		\label{fig:model}
	\end{figure}

    As shown in Figure~\ref{fig:model}, the model receives word representations and produces simultaneously a representation for each pair in the sentence. 
    These representations combine the target pair, its context words, their relative positions to the pair entities and walks between them.
    During classification they are used to predict the relation type of each pair. 
	
	\subsection{Embedding Layer}
    
	The embedding layer involves the creation of $n_w$, $n_t$, $n_p$-dimensional vectors which are assigned to words, semantic entity types and relative positions to the target pairs. 
	We map all words and semantic types into real-valued vectors $\mathbf{w}$ and $\mathbf{t}$ respectively. 
	Relative positions to target entities are created based on the position of words in the sentence. 
	In the example of Figure~\ref{fig:ex}, the relative position of \textit{teammates} to \textit{capital} is $-3$ and the relative position of \textit{teammates} to \textit{Toefting} is $+16$. 
	We embed real-valued vectors $\mathbf{p}$ to these positions.

	\subsection{Bidirectional LSTM Layer}
    
	The word representations of each sentence are fed into a Bidirectional Long-short Term Memory (BLSTM) layer, which encodes the context representation for every word. 
	The BLSTM outputs new word-level representations $\mathbf{h}$~\cite{hochreiter1997long} that consider the sequence of words.
    
	We avoid encoding target pair-dependent information in this BLSTM layer. 
    This has two advantages: 
	(i) the computational cost is reduced as this computation is repeated based on the number of sentences instead of the number of pairs, 
	(ii) we can share the sequence layer among the pairs of a sentence. 
	The second advantage is particularly important as it enables the model to indirectly learn hidden dependencies between the related pairs in the same sentence. 
    
	For each word $t$ in the sentence, we concatenate the two representations from left-to-right and right-to-left pass of the LSTM into a $n_e$-dimensional vector, $\mathbf{e}_t = {[ \overrightarrow{\mathbf{h}_t}; \overleftarrow{\mathbf{h}_t} ]}$.
	
	\subsection{Edge Representation Layer}
    
	The output word representations of the BLSTM are further divided into two parts: (i) target pair representations and (ii) target pair-specific context representations. 
	The context of a target pair can be expressed as all words in the sentence that are not part of the entity mentions. 
	We represent a related pair as described below.

	A target pair contains two entities $e_i$ and $e_j$. 
    If an entity consists of $N$ words, we create its BLSTM representation as the average of the BLSTM representations of the corresponding words, 
    $\mathbf{e} = \frac{1}{|I|}\sum_{i \in I} {\mathbf{e}_i}$, where $I$ is a set with the word indices inside entity $e$.
    
    We first create a representation for each pair entity and then we construct the representation for the context of the pair.
     The representation of an entity $e_i$ is the concatenation of its BLSTM representation $\mathbf{e}_i$, the representation of its entity type $\mathbf{t}_i$ and the representation of its relative position to entity $e_j$, $\mathbf{p}_{ij}$. 
     Similarly, for entity $e_j$ we use its relative position to entity $e_i$, $\mathbf{p}_{ji}$.
     Finally, the representations of the pair entities are as follows:
     $\mathbf{v}_i = {[ \mathbf{e}_i ; \mathbf{t}_i ; \mathbf{p}_{ij} ]}$ and 
	 $\mathbf{v}_j = {[ \mathbf{e}_j ; \mathbf{t}_j ; \mathbf{p}_{ji} ]}$. 
    
	The next step involves the construction of the representation of the context for this pair.
    For each context word $w_z$ of the target pair $e_i$, $e_j$, we concatenate its BLSTM representation $\mathbf{e}_z$, its semantic type representation $\mathbf{t}_z$ and two relative position representations: to target entity $e_i$, $\mathbf{p}_{zi}$ and to target entity $e_j$, $\mathbf{p}_{zj}$.
    The final representation for a context word $w_z$ of a target pair is,
    $\mathbf{v}_{ijz} = [ \mathbf{e}_z ; \mathbf{t}_z ; \mathbf{p}_{zi} ; \mathbf{p}_{zj} ]$.
    For a sentence, the context representations for all entity pairs can be expressed as a three-dimensional matrix $\mathbf{C}$, where rows and columns correspond to entities and the depth corresponds to the context words.
    
	The context words representations of each target pair are then compiled into a single representation with an attention mechanism. 
	Following the method proposed in \citet{zhou2016attention}, we calculate weights for the context words of the target-pair and compute their weighted average,
	\begin{equation}
	\begin{aligned}
	\mathbf{u}	  	&=  \mathbf{q}^{\top} \; \tanh(\mathbf{C}_{ij}) , \\
	\mathbf{\alpha} &=  \softmax(\mathbf{u}) , 					      \\
	\mathbf{c}_{ij} &=  \mathbf{C}_{ij} \; \mathbf{\alpha}^{\top},
	\end{aligned}
	\label{eq:att}
	\end{equation}
	where $\mathbf{q} \in \mathbb{R}^{n_d}, n_d = n_e + n_t + 2n_p $ denotes a trainable attention vector, 
	$\alpha$ is the attended weights vector
	and
	$\mathbf{c}_{ij} \in \mathbb{R}^{n_d}$ is the context representation of the pair as resulted by the weighted average.
    This attention mechanism is independent of the relation type. We leave relation-dependent attention as future work.
	
    Finally, we concatenate the representations of the target entities and their context ($ \in \mathbb{R}^{n_m}$). We use a fully connected linear layer, $\mathbf{W}_s \in \mathbb{R}^{n_m \times n_s}$ with $n_s < n_m$ to reduce the dimensionality of the resulting vector. 
    This corresponds to the representation of an edge or a one-length walk between nodes $i$ and $j$:
	$\mathbf{v}^{(1)}_{ij} = \mathbf{W}_s \; [ \mathbf{v}_{i} ; \mathbf{v}_{j} ; \mathbf{c}_{ij} ] \in \mathbb{R}^{n_s}$. 
	
	\subsection{Walk Aggregation Layer}
    
    Our main aim is to support the relation between an entity pair by using chains of intermediate relations between the pair entities.
    Thus, the goal of this layer is to generate a single representation for a finite number of different lengths walks between two target entities.
	To achieve this, we represent a sentence as a directed graph, where the entities constitute the graph nodes and edges correspond to the representation of the relation between the two nodes. 
	The representation of one-length walk between a target pair $\mathbf{v}^{(1)}_{ij}$, serves as a building block in order to create and aggregate representations for one-to-$l$-length walks between the pair. 
	The walk-based algorithm can be seen as a two-step process: walk construction and walk aggregation.
	During the first step, two consecutive edges in the graph are combined using a modified bilinear transformation, 
	\begin{equation}
    f(\mathbf{v}^{(\lambda)}_{ik}, \mathbf{v}^{(\lambda)}_{kj}) = \sigma \left( \mathbf{v}^{(\lambda)}_{ik} \; \odot \; ( \mathbf{W}_b \; \mathbf{v}^{(\lambda)}_{kj} ) \right),
	\label{eq:w_construct}
	\end{equation}
	where $\mathbf{v}^{(\lambda)}_{ij} \in \mathbb{R}^{n_b}$ corresponds to walks representation of lengths one-to-$\lambda$ between entities $e_i$ and $e_j$,
    $\odot$ represents element-wise multiplication,
	$\sigma$ is the sigmoid non-linear function and 
	$\mathbf{W}_b \in \mathbb{R}^{n_b \times n_b}$ is a trainable weight matrix. 
    This equation results in walks of lengths two-to-2$\lambda$.
	
	In the walk aggregation step, we linearly combine the initial walks (length one-to-$\lambda$) and the extended walks (length two-to-$2\lambda$),
	\begin{equation}
	\mathbf{v}_{ij}^{(2\lambda)} = \beta \mathbf{v}_{ij}^{(\lambda)} + (1 - \beta) \sum_{k \neq i,j} f(\mathbf{v}_{ik}^{(\lambda)}, \mathbf{v}_{kj}^{(\lambda)}),
	\label{eq:w_aggregate}
	\end{equation}
	where 
	$\beta$ is a weight that indicates the importance of the shorter walks.
    Overall, we create a representation for walks of length one-to-two using Equation (\ref{eq:w_aggregate}) and $\lambda=1$. We then create a representation for walks of length one-to-four by re-applying the equation with $\lambda=2$.   
	We repeat this process until the desired maximum walk length is reached, which is equivalent to $2\lambda = l$.
    
	\subsection{Classification Layer}
    
	For the final layer of the network, we pass the resulted pair representation into a fully connected layer with a softmax function,
	\begin{equation}
	\mathbf{y} = \softmax(\mathbf{W}_r \mathbf{v}_{ij}^{(l)} + \mathbf{b}_r),
	\end{equation}
	where $\mathbf{W}_r \in \mathbb{R}^{n_b \times n_r}$ is the weight matrix, 
	$n_r$ is the total number of relation types  
	and $b_r$ is the bias vector.
	
	We use in total $2r + 1$ classes in order to consider both directions for every pair, i.e., left-to-right and right-to-left. The first argument appears first in a sentence in a left-to-right relation while the second argument appears first in a right-to-left relation. The additional class corresponds to non-related pairs, namely ``no relation" class.
	We choose the most confident prediction for each direction and choose the positive and most confident prediction when the predictions contradict each other.

	\section{Experiments}
	
	\subsection{Dataset}
    
	We evaluate the performance of our model on ACE 2005\footnote{\url{https://catalog.ldc.upenn.edu/ldc2006t06}} for the task of relation extraction.
	ACE 2005 includes $7$ entity types and $6$ relation types between named entities.  
	We follow the preprocessing described in \citet{miwa2016end}.

	\subsection{Experimental Settings}
    
	We implemented our model using the Chainer library~\citep{tokui2015chainer}.\footnote{\url{https://chainer.org/}} 
	The model was trained with Adam optimizer~\citep{kingma2014adam}. 
	We initialized the word representations with existing pre-trained embeddings with dimensionality of $200$.\footnote{\url{https://github.com/tticoin/LSTM-ER}}
    Our model did not use any external tools except these embeddings.
    
	The forget bias of the LSTM layer was initialized with a value equal to one following the work of \citet{jozefowicz2015empirical}. 
	We use a batchsize of $10$ sentences and fix the pair representation dimensionality to $100$. We use gradient clipping, dropout on the embedding and output layers and L2 regularization without regularizing the biases, to avoid overfitting. We also incorporate early stopping with patience equal to five, to chose the number of training epochs and parameter averaging.
	We tune the model hyper-parameters on the respective development set using the RoBO Toolkit~\cite{klein-bayesopt17}. 
	Please refer to the supplementary material for the values.
	
    We extract all possible pairs in a sentence based on the number of entities it contains. 
    If a pair is not found in the corpus, it is assigned the ``no relation'' class. 	
    We report the micro precision, recall and F1 score following~\citet{miwa2016end} and~\citet{nguyen2015perspective}.

	\section{Results}
     
	Table~\ref{tab:res} illustrates the performance of our proposed model in comparison with \textit{SPTree} system~\citet{miwa2016end} on ACE 2005.  
    We use the same data split with \textit{SPTree} to compare with their model. We retrained their model with gold entities in order to compare the performances on the relation extraction task. 
    The \textit{Baseline} corresponds to a model that classifies relations by using only the representations of entities in a target pair.
     
	\begin{table}[t!]
		\centering
		\renewcommand*{\arraystretch}{1.1}
		\begin{tabular}{|lccc|}
			\hline
			Model & P & R & F1 (\%) \\
			\hline \hline 
			SPTree				    & 70.1 & 61.2 & 65.3  \\
			\hline
			Baseline  			    & 72.5 & 53.3 & 61.4\rlap{$^*$}  \\
			No walks $l$ = 1	    & 71.9 & 55.6 & 62.7  \\
			\ \ + Walks $l$ = 2	    & 69.9 & 58.4 & 63.6\rlap{$^\diamond$}  \\
			\ \ + Walks $l$ = 4 	& 69.7 & 59.5 & 64.2\rlap{$^\diamond$}  \\
            \ \ + Walks $l$ = 8 	& 71.5 & 55.3 & 62.4  \\
			\hline
		\end{tabular}
		\caption{Relation extraction performance on ACE 2005 test dataset. * denotes significance at p $<$ 0.05 compared to SPTree, $\diamond$ denotes significance at p $<$ 0.05 compared to the Baseline. } 
        \label{tab:res}
	\end{table}
    
    As it can be observed from the table, the \textit{Baseline} model achieves the lowest F1 score between the proposed models. 
    By incorporating attention we can further improve the performance by 1.3 percent point (pp). 
    The addition of $2$-length walks further improves performance (0.9 pp).
    The best results among the proposed models are achieved for maximum $4$-length walks.
    By using up-to $8$-length walks the performance drops almost by 2 pp.  
    We also compared our performance with \citet{nguyen2015perspective} (\textit{CNN}) using their data split.\footnote{The authors kindly provided us with the data split.} For the comparison, we applied our best performing model ($l$ = 4).\footnote{We kept the same parameters when we apply our model to the this data split. We did not remove any negative examples unlike the \textit{CNN} model.}
    The obtained performance is 65.8 / 58.4 / 61.9 in terms of P / R / F1 (\%) respectively. 
    In comparison with the performance of the \textit{CNN} model, 71.5 / 53.9 / 61.3, we observe a large improvement in recall which results in 0.6 pp F1 increase.
    
    We performed the Approximate Randomization test~\cite{noreen1989computer} on the results. The best walks model has no statistically significant difference with the state-of-the-art \textit{SPTree} model as in Table~\ref{tab:res}. This indicates that the proposed model can achieve comparable performance without any external syntactic tools.
    
    \begin{table}[t!]
		\centering
		\renewcommand*{\arraystretch}{1.1}
		\begin{tabular}{|lcccc|}
			\hline
			\# Entities & $l = 1$ & $l$ = 2 & $l$ = 4 & $l$ = 8 \\
			\hline \hline 
 			$2$		    & 71.2 & 69.8 			 & 72.9            & 71.0 \\
 			$3$ 	    & 70.1 & 67.5 			 & 67.8            & 63.5\rlap{$^*$} \\ 
            $[4, 6)$  	& 56.5 & 59.7 			 & 59.3            & 59.9 \\
            $[6, 12)$   & 59.2 & 64.2\rlap{$^*$} & 62.2            & 60.4 \\
            $[12, 23)$  & 54.7 & 59.3            & 62.3\rlap{$^*$} & 55.0 \\
			\hline
		\end{tabular}
		\caption{Relation extraction performance (F1 \%) on ACE 2005 development set for different number of entities. * denotes significance at p $<$ 0.05 compared to $l = 1$.} 
        \label{tab:anal}
	\end{table}
    
    Finally, we show the performance of the proposed model as a function of the number of entities in a sentence. Results in Table~\ref{tab:anal} reveal that for multi-pair sentences the model performs significantly better compared to the no-walks models, proving the effectiveness of the method. Additionally, it is observed that for more entity pairs, longer walks seem to be required. However, very long walks result to reduced performance ($l$ = 8).

	\section{Related Work}
	Traditionally, relation extraction approaches have incorporated a large variety of hand-crafted features to represent related entity pairs \cite{hermann2013role,miwa2014table,nguyen2014employing,gormley2015embedding}. 
	Recent models instead employ neural network architectures and achieve state-of-the-art results without heavy feature engineering. 
	Neural network techniques can be categorized into recurrent neural networks (RNNs) and convolutional neural networks (CNNs). The former is able to encode linguistic and syntactic properties of long word sequences, making them preferable for sequence-related tasks, e.g. natural language generation~\cite{goyal2016natural}, machine translation~\cite{sutskever2014sequence}. 
	
	State-of-the-art systems have proved to achieve good performance on relation extraction using RNNs \cite{cai2016bidirectional,miwa2016end,xu2016improved,liu2015dependency}. Nevertheless, most approaches do not take into consideration the dependencies between relations in a single sentence \cite{santos2015ranking,nguyen2015perspective} and treat each pair separately.
	Current graph-based models are applied on knowledge graphs for distantly supervised relation extraction~\cite{zeng2017paths}. Graphs are defined on semantic types in their method, whereas we built entity-based graphs in sentences. 
    Other approaches also treat multiple relations in a sentence~\cite{gupta2016table,miwa2014table,li2014incremental}, but they fail to model long walks between entity mentions.

	\section{Conclusions}  
	We proposed a novel neural network model for simultaneous sentence-level extraction of related pairs. 
    Our model exploits target and context pair-specific representations and creates pair representations that encode up-to $l$-length walks between the entities of the pair. 
    We compared our model with the state-of-the-art models and observed comparable performance on the ACE2005 dataset without any external syntactic tools. 
    The characteristics of the proposed approach are summarized in three factors: the encoding of dependencies between relations, the ability to represent multiple walks in the form of vectors and the independence from external tools. 
    Future work will aim at the construction of an end-to-end relation extraction system as well as application to different types of datasets.

	\section*{Acknowledgments}
This research has been carried out with funding from AIRC/AIST, the James Elson Studentship Award, BBSRC grant BB/P025684/1 and MRC MR/N00583X/1. Results were obtained from a project commissioned by the New Energy and Industrial Technology Development Organization (NEDO). We would finally like to thank the anonymous reviewers for their helpful comments.

	\bibliography{graphRE_biblio}
	\bibliographystyle{acl_natbib}
	
    \appendix
    
    \section{Hyper-parameter Settings}
    We tuned our proposed model using the RoBO toolkit (\url{https://github.com/automl/RoBO}). 
    Table~\ref{tab:opt} provides the selected options we used for tuning the model.
	\begin{table}[ht!]
		\centering
		\renewcommand*{\arraystretch}{1.1}
			\begin{tabular}{|l|r|}
				\hline
				\multicolumn{2}{|c|}{Optimization Options} \\
				\hline \hline
				Optimization method  & Bohamiann \\
				Maximizer 			 & scipy \\
				Acquisition function & log\_ei \\
				Number of iterations & $50$ \\
				Initial points 		 & $3$ \\
				\hline
			\end{tabular}
			\caption{Hyper-parameters optimization options.}
			\label{tab:opt}
		\end{table}
        
	The parameters that gave the best performance for the different models can be found in Tables~\ref{tab:params_a}-\ref{tab:params_e}.
	\begin{table}[ht!]
    		\centering
			\begin{subtable}[ht!]{0.4\textwidth}
	             \begin{tabular}{|l|r|}
	                 \hline 
	                 Parameter & Baseline \\
	                 \hline \hline
	                 Position dimension $n_p$  & $25$ \\
	                 Type dimension $n_t$	  & $15$ \\
	                 LSTM dimension $n_e$	  & $100$ \\
	                 Input layer dropout 	  & $0.3$ \\
	                 Output layer dropout 	  & $0.03$ \\
	                 Learning rate 			  & $0.0018$ \\
	                 Regularization 		  & $3.2 \cdot 10^{-5}$  \\
	                 Gradient clipping		  & $25.63$ \\
	                 \hline
	             \end{tabular}
             	\subcaption{}
             	\label{tab:params_a}
			\end{subtable}
	             
            \vspace{0.5cm}
            
			\begin{subtable}[ht!]{0.4\textwidth}
	             \begin{tabular}{|l|r|}
	                 \hline 
	                 Parameter & $l$ = 1 \\
	                 \hline \hline
	                 Position dimension $n_p$  & $25$ \\
	                 Type dimension $n_t$	  & $25$ \\
	                 LSTM dimension $n_e$	  & $100$ \\
	                 Input layer dropout 	  & $0.13$ \\
	                 Output layer dropout 	  & $0.38$ \\
	                 Learning rate 			  & $0.0017$ \\
	                 Regularization 		  & $6.1 \cdot 10^{-5}$  \\
	                 Gradient clipping		  & $30$ \\
	                 \hline
	             \end{tabular}
             	\caption{}
             	\label{tab:params_b}
			\end{subtable}
	             
            \vspace{0.5cm}
            
			\begin{subtable}[ht!]{0.4\textwidth}
	             \begin{tabular}{|l|r|}
	                 \hline 
	                 Parameter & $l$ = 2 \\
	                 \hline \hline
	                 Position dimension $n_p$  & $25$ \\
	                 Type dimension $n_t$	  & $20$ \\
	                 LSTM dimension $n_e$	  & $100$ \\
	                 $\beta$				  & $0.72$ \\
	                 Input layer dropout 	  & $0.25$ \\
	                 Output layer dropout 	  & $0.37$ \\
	                 Learning rate 			  & $0.003$ \\
	                 Regularization 		  & $0.0001$  \\
	                 Gradient clipping		  & $8.6$ \\
	                 \hline
	             \end{tabular}
             	\subcaption{}
             	\label{tab:params_c}
			\end{subtable}
            
            \vspace{0.5cm}
            
            \begin{subtable}[ht!]{0.4\textwidth}
	             \begin{tabular}{|l|r|}
	                 \hline 
	                 Parameter & $l$ = 4 \\
	                 \hline \hline
	                 Position dimension $n_p$  & $25$ \\
	                 Type dimension $n_t$	  & $20$ \\
	                 LSTM dimension $n_e$	  & $100$ \\
	                 $\beta$				  & $0.77$ \\
	                 Input layer dropout 	  & $0.11$ \\
	                 Output layer dropout 	  & $0.32$ \\
	                 Learning rate 			  & $0.002$ \\
	                 Regularization 		  & $5.7 \cdot 10^{-5}$  \\
	                 Gradient clipping		  & $24.4$ \\
	                 \hline
	             \end{tabular}
             	\subcaption{}
             	\label{tab:params_d}
			\end{subtable}
	  \end{table}  
	
    \vspace{0.5cm}
    
	\begin{table}[ht!]
		\ContinuedFloat 
			\begin{subtable}[h]{0.4\textwidth}
	             \begin{tabular}{|l|r|}
	                 \hline 
	                 Parameter & $l$ = 8 \\
	                 \hline \hline
	                 Position dimension $n_p$  & $25$ \\
	                 Type dimension $n_t$	  & $20$ \\
	                 LSTM dimension $n_e$	  & $100$ \\
	                 $\beta$				  & $0.88$ \\
	                 Input layer dropout 	  & $0.49$ \\
	                 Output layer dropout 	  & $0.36$ \\
	                 Learning rate 			  & $0.001$ \\
	                 Regularization 		  & $1.88 \cdot 10^{-5}$  \\
	                 Gradient clipping		  & $10.5$ \\
	                 \hline
	             \end{tabular}
             	 \subcaption{}
             	 \label{tab:params_e}
 			\end{subtable}
            
 		\caption{Best hyper-parameters settings for proposed models.}
 		\label{tab:params}
	\end{table}
        
\end{document}